\pdfoutput=1 
\documentclass[sigconf,natbib=true]{acmart}

\usepackage{geometry}

\usepackage{multirow}
\usepackage{caption}

\usepackage{amssymb}
\usepackage{amsmath}
\usepackage{subcaption}
\usepackage{balance}
\usepackage{url}
\usepackage[normalem]{ulem}
\usepackage{array}
\usepackage{makecell}
\usepackage{pifont}

\newcommand\ies{\textit{i.e.}}
\newcommand\egs{\textit{e.g.}}

\newcommand\figcaption{\def\@captype{figure}\caption}
\newcommand\tabcaption{\def\@captype{table}\caption}

\newcommand\bs[1]{\mathbf{#1}}

\usepackage{algorithm}
\usepackage{algorithmic}

\AtBeginDocument{%
	\providecommand\BibTeX{{%
			\normalfont B\kern-0.5em{\scshape i\kern-0.25em b}\kern-0.8em\TeX}}}

\copyrightyear{2023} 
\acmYear{2023} 
\acmPrice{15.00}
\acmDOI{XXXXX/XXXXXXXXXXX}
\acmISBN{XXXXXXXXXXXXXX}
\author{Jiabang He}
\affiliation{%
	\institution{Center for Future Media \& School of Computer Science and Engineering}
	\institution{University of Electronic Science and Technology of China}
	\city{Chengdu}
	\country{China}}

\author{Yi Hu}
\affiliation{%
	\institution{Center for Future Media \& School of Computer Science and Engineering}
	\institution{University of Electronic Science and Technology of China}
	\city{Chengdu}
	\country{China}}

\author{Lei Wang}
\authornote{Corresponding author.}
\affiliation{%
	\institution{Singapore Management University}
	\country{Singapore}}

\author{Xing Xu}
\authornote{Corresponding author.}
\affiliation{%
	\institution{Center for Future Media \& School of Computer Science and Engineering}
	\institution{University of Electronic Science and Technology of China}
	\city{Chengdu}
	\country{China}}

\author{Ning Liu}
\affiliation{%
	\institution{Beijing Forestry University}
	\city{Beijing}
	\country{China}}

\author{Hui	Liu}
\affiliation{%
	\institution{Beijing Rongda Technology Co., Ltd.}
	\city{Beijing}
	\country{China}}

\author{Heng Tao Shen}
\affiliation{%
	\institution{Center for Future Media \& School of Computer Science and Engineering}
	\institution{University of Electronic Science and Technology of China}
	\city{Chengdu}
	\country{China} \\
	\institution{Peng Cheng Laboratory}
	\city{Shenzhen}
	\country{China}
}

\begin{document}
\fancyhead{}
	\title{
		Do-GOOD: Towards Distribution Shift Evaluation for Pre-Trained Visual Document Understanding Models
	}
	
	
 
	\begin{abstract}
		Numerous pre-training techniques for visual document understanding (VDU) have recently shown substantial improvements in performance across a wide range of document tasks. 
		However, these pre-trained VDU models cannot guarantee continued success when the distribution of test data differs from the distribution of training data. 
		In this paper, to investigate how robust existing pre-trained VDU models are to various distribution shifts, we first develop an \textbf{o}ut-\textbf{o}f-\textbf{d}istribution (OOD) benchmark termed Do-GOOD for the fine-\textbf{G}rained analysis on \textbf{Do}cument image-related tasks specifically.
		The Do-GOOD benchmark defines the underlying mechanisms that result in different distribution shifts and contains 9 OOD datasets covering 3 VDU related tasks, \egs, document information extraction, classification and question answering.
		We then evaluate the robustness and perform a fine-grained analysis of 5 latest VDU pre-trained models and 2 typical OOD generalization algorithms on these OOD datasets.
		Results from the experiments demonstrate that there is a significant performance gap between the in-distribution (ID) and OOD settings for document images, and that fine-grained analysis of distribution shifts can reveal the brittle nature of existing pre-trained VDU models and OOD generalization algorithms.
		The code and datasets for our Do-GOOD benchmark can be found at \color{gray}{\url{https://github.com/MAEHCM/Do-GOOD}}.
	\end{abstract}

	
	
	
	
	\maketitle
	
	\section{Introduction}
	\label{sec:intro}
	
	\begin{figure}[!htb]
		\centering
		\includegraphics[width=1.0\linewidth]{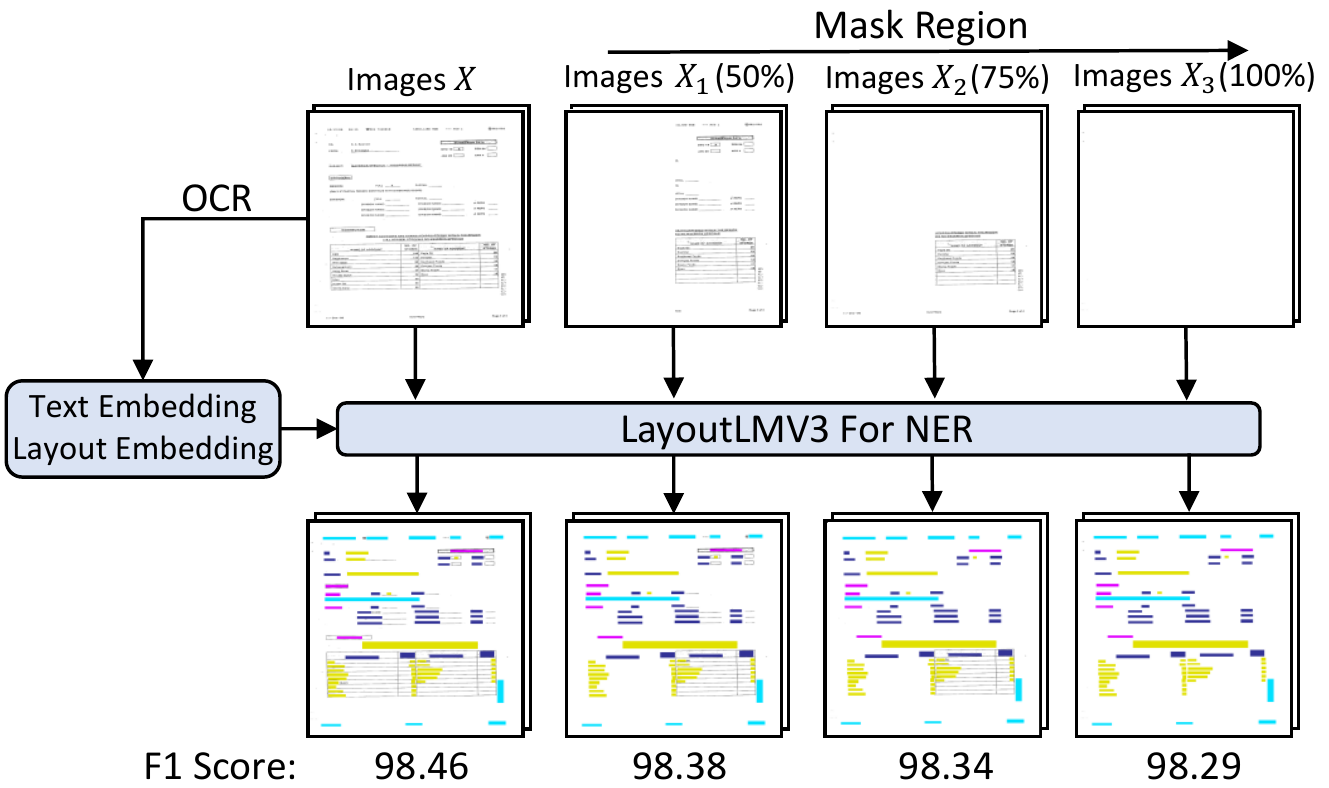}
		\caption{
			Illustration of the different importance of the input image, text, and layout embedding for the LayoutLMV3~\cite{huang2022layoutlmv3} model on the document Named Entity Recognition (NER) task. 
			Notably, though the image $x$ is masked with different proportions, \ies~ $50\%$, $75\%$ and $100\%$, the model prediction (F1 score) just slightly changes.
		}
		\label{fig:cases}
	\end{figure}
	
	\noindent\textbf{Background}. Document images (e.g., invoices and lease agreements), typically containing rich contextual text and structural information, are commonly seen in modern working and living environments.
	Automatic processing and understanding of document images have wide-ranging use cases in real-world scenarios, such as document image classification~\cite{harley2015icdar, harley2015evaluation, xu2020layoutlm}, information extraction from document images~\cite{park2019cord, jaume2019funsd, majumder2020representation}, and document visual question answering~\cite{mathew2021docvqa}.
	Recently, numerous pre-training techniques concerning document image understanding have been proposed and shown to be effective for various document tasks~\cite{kim2022donut, huang2022layoutlmv3, li2021selfdoc, xu2020layoutlm, gu2022xylayoutlm, wang2022LiLT, li2022dit}.
	Despite the encouraging results achieved by these models, it cannot be guaranteed that models designed under the same training and test data distribution would continue to perform well when the distribution of test data differs from the training data distribution ~\cite{larson2022evaluating, cao2022query}. 
	\textit{However, most document datasets~\cite{harley2015icdar, park2019cord, jaume2019funsd, mathew2021docvqa} are designed following the i.i.d. assumption, with the training and test data from the same distribution.} 
	
	\textbf{Motivation}. To enable the models for document classification to have the ability to handle \textit{out-of-distribution} (OOD) document images, Larson et al.~\cite{larson2022evaluating} present a new OOD testbed in terms of a widely-used document classification benchmark dataset namely RVL-CDIP. 
	This RVL-CDIP OOD benchmark is only used to develop and evaluate the robustness of methods for document image classification, which just need the models to have the capacity to model coarse-grained information over document images.

	Although the RVL-CDIP OOD benchmark reveals that image information is quite important for document classification, image information \textit{plays a relatively minor role on other document imaging tasks}, such as information extraction~\cite{zhang2022end, ru2020quachie}. 
	As illustrated in Figure~\ref{fig:cases}, taking the NER task on the latest pre-trained visual document understanding (VDU) model LayoutLMv3 \cite{huang2022layoutlmv3} for example, when different proportions of an input image $x$ are masked as blank, the F1 score of the LayoutLMv3 model prediction is basically the same. 
	It indicates the prediction of the LayoutLMv3 model relies more on the text and layout information rather than the visual cues.
	Besides, document images naturally possess three distinct features, including image, text, and layout information. The tasks, such as information extraction from document images and document visual question answering, necessitate a fine-grained understanding of complicated interactions over image, text, and layout information~\cite{jaume2019funsd, park2019cord, mathew2021docvqa}.
	On the other hand, models designed based on these three types of features require image, text, and layout modules to carry different perspectives of input information for a document image~\cite{xu2020layoutlm, xu-etal-2021-layoutlmv2, huang2022layoutlmv3}.
	The uniqueness of document image data calls for the construction of document image specific OOD benchmarks with various distribution shifts.
	This naturally begs the following question: 
	\textit{How robust are existing pre-trained VDU  models to fine-grained distribution shifts occurring on document image tasks?}
	
	\begin{figure*}[!htb]
		\centering
		\includegraphics[width=0.9\linewidth]{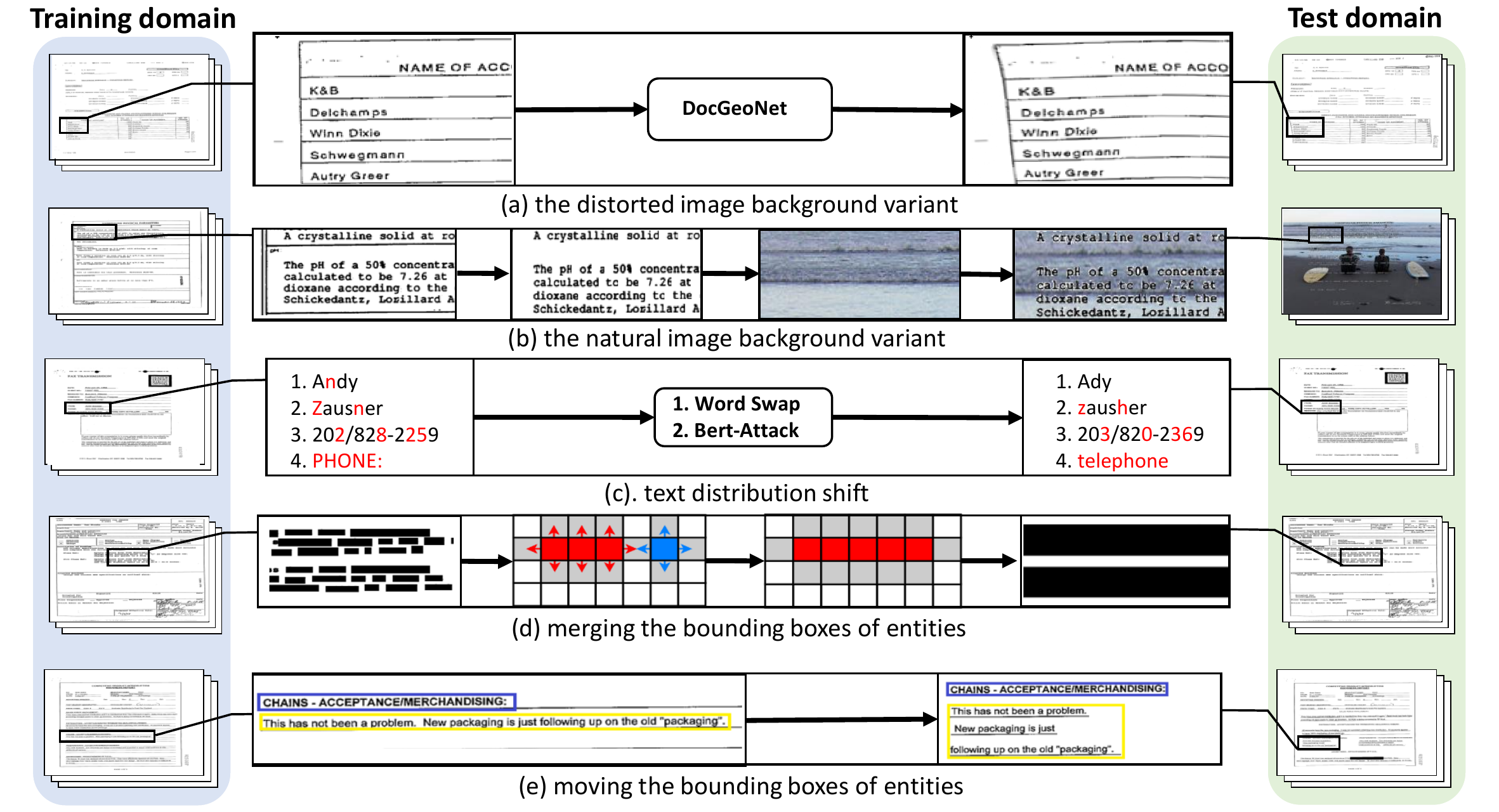}
		\caption{
			In the Do-GOOD benchmark, each document image is extracted from the training domain. We studied five distribution shifts acting on the three modalities respectively to generate the test domain. 
			The five distribution shift acting includes: two image distribution shifts with (a) the distorted image background or (b) the natural image background; (c) text distribution shift with Bert-Attack and Word Swap; two layout distribution shifts by (d) merging and (e) moving the layouts.
		}
		\label{fig:cases2}
	\end{figure*}
	
	\textbf{Contribution}. To answer the above question regarding the robust estimation of the VDU models' capability in the document image OOD scenario, in this paper, we aim to develop a systematic document image OOD benchmark, namely Do-GOOD. 
	To design Do-GOOD, we adhere to the following criteria. 
	In particular, we expect that (1) A large distribution gap between training and test data can result in a substantial drop in model performance; (2) Fine-grained analysis of distribution shifts can expose the brittle nature of existing models; (3) Designed benchmark datasets should be possibly solvable, easily scalable, and human-readable.
	
	To meet criteria (2), as shown in Figure~\ref{fig:cases2}, we divide distribution shifts into three categories of different characteristics, i.e., image, text, and layout distribution shifts. 
	The distribution shifts are used to examine the partiality of VDU models on text, image, and layout information, which could compromise the robustness of VDU models.
	For image shift, we first disentangle the content (e.g., text on form images) from the background (e.g., table borders on form images) and then replace the background with a natural image from MSCOCO.
	For text shift, we employ common text attacks, such as BERT-Attack~\cite{Bert-attack:li2020bert} and Word Swap~\cite{morris2020textattack, mrkvsic2016counter}, to simulate a more realistic scenario where input document images may contain problematic text caused by OCR errors.
	We have two strategies to induce layout shifts. The first involves merging smaller bounding boxes to form a larger box. Another option is to move a particular box to a different location on the document image.
	These carefully designed strategies from image, text, and layout perspectives can automatically produce OOD testbeds having substantially different distributions from the training distributions, thus meeting criteria (1) and (2).
	
	Here is a summary of our main contributions:
	(1) We provide a fine-grained analysis of various distribution shifts in document images from image, text, and layout perspectives;
	(2) To generate the OOD benchmark that meets the aforementioned three criteria, we introduce a suite of automatic strategies to generate OOD data;
	(3) We evaluate and compare 5 state-of-the-art pre-trained VDU models and 2 representative OOD algorithms in generated OOD testbeds (i.e., Do-GOOD) across different document image tasks.
	We hope that the proposed Do-GOOD benchmark, the empirical study, and our in-depth analysis will benefit future research to improve the robustness of pre-trained VDU models.

	\section{Related Work}
	\label{sec:rel}
	
	\noindent\textbf{Visual Document Understanding}.
	Visual document classification~\cite{harley2015icdar}, visual document information extraction~\cite{park2019cord, jaume2019funsd}, and visual question answering on documents~\cite{mathew2021docvqa} are among the core tasks of automated document processing. 
	For visual document classification, early works model visual information by various CNN-based methods~\cite{harley2015evaluation, kang2014convolutional}.
	Based on the output of OCR, RNN-based~\cite{sage2019recurrent} and Transformer-based~\cite{majumder2020representation} models predict the label for the text.
	In DocVQA, an LSTM encoder is used to model textual information and CNN encoders are used to model visual information to answer questions about document images~\cite{mathew2021docvqa}.
	As a result of the recent success of large-scale pre-training in NLP, such as BERT~\cite{devlin2019bert} and RoBERTa~\cite{liu2019roberta}, most methods take pre-train-and-fine-tune schemes for addressing downstream tasks together~\cite{xu2020layoutlm,xu-etal-2021-layoutlmv2,garncarek2021lambert,Powalski2021GoingFB,wu2021lampret,Li2021StructuralLMSP,li2021selfdoc,li2021structext,wang2022LiLT,gu2022xylayoutlm, huang2022layoutlmv3, li2022dit, kim2022donut}.
	
	The majority of current state-of-the-art models separate scanned document images into text, vision, and layout attributes and design modules to process them individually or together.
	For example, most methods begin by obtaining text tokens and layouts from OCR tools and then feed the OCR-ed text into pre-trained language models to model the text information.
	For extracting region features, some works use object detectors~\cite{xu2020layoutlm,Powalski2021GoingFB,li2021selfdoc,gu2021unidoc}, while others use the vision transformer~\cite{kim2021vilt, li2022dit, kim2022donut, huang2022layoutlmv3}.
	LayoutLM~\cite{xu2020layoutlm} and its followings~\cite{xu-etal-2021-layoutlmv2, xu2021layoutxlm} employ two-dimensional positional vectors for the layout information and fuse their transformed vectors with text embeddings for the multimodal pre-trained model.
	After collecting text, image, and layout features, most methods leverage a multimodal fusion module to encourage modeling interactions between them.
	For some exceptions, Donut~\cite{kim2022donut} conducts inference in an end-to-end fashion without OCR processing.
	LayoutLMv3~\cite{huang2022layoutlmv3} makes use of patch-level embeddings for text and image patches for alignment on document images.

	\noindent\textbf{Out-of-Distribution Benchmarks}.
	We briefly review benchmarks for distribution shifts in this section.
	Distribution shifts have been a long-standing problem in the machine learning community~\cite{MORENOTORRES2012521, shen2021towards}.
	Recently, increasing research has shifted their attention from achieving the highest performance under in-distribution (ID) settings towards assessing models' robustness and generalization capacities~\cite{zhang2017mixup, sun2016deep, sagawa2019distributionally, peters2016causal, krueger2021out, ganin2016domain, arjovsky2019invariant, ahuja2021invariance}.
	To this end, various OOD benchmarks have been created to encourage the building of more robust models~\cite{gulrajani2020search, he2021towards, koh2021wilds, ye2021ood, gui2022good}.
	WILDS~\cite{koh2021wilds} creates a curated benchmark of 10 datasets ranging from the categorization of animal species to code completion. This benchmark requires curated datasets that express large distribution shifts, are relevant in the real world, and can potentially be solved.
	The GOOD benchmark~\cite{gui2022good} is designed to graph OOD method evaluations based on two shifts.
	Beyond, Wiles et al.~\cite{wiles2021fine} provide a holistic analysis of current SOTA methods by evaluating multiple distinct methods across both synthetic and real-world datasets.
	
	There are also OOD benchmarks for document image tasks.
	Larson et al.~\cite{larson2022evaluating} establish an OOD testbed comprised of RVL-CDIP-N and RVL-CDIP-O. RVL-CDIP-N consists of in-domain documents sampled from a different distribution than RVL-CDIP. 
	VL-CDIP-O comprises out-of-domain document images that do not fall into RVL-CDIP categories.
	The LastDoc4000~\cite{cao2022query} is designed for situations in which input document images may contain unknown layouts and keys caused by OCR errors.
	However, the existing two benchmarks either ignore layout or text distribution shifts and only focus on document IE tasks.
	In contrast to them, Do-GOOD considers distribution shifts of text, vision, and layout across multiple common document image tasks from image-centric to text-centric perspectives.

	\section{Do-GOOD Benchmark Design}
	\label{sec:ben}
	Existing datasets, such as FUNSD~\cite{jaume2019funsd}, prepare training and test samples under the i.i.d. assumption.
	Given a data distribution $p_{\text{train}}$ of training inputs $x$, the goal of a document image model $f$ is to minimize the risk $R$ as follows:
	\begin{equation}
		R(f)=\mathbb{E}_{\left(\boldsymbol{x}, y^l\right) \sim p_{\text{train}}}\left[\mathcal{L}\left(y^l, f(\boldsymbol{x})\right)\right],
	\end{equation}
	where $\mathcal{L}$ is the loss function for a particular task.
	Due to confounding factors, such as selection bias in the data collection process and random data splits, it is difficult for train and test data to follow the same data distribution in practice (i.e., $p_{\text {train}}\neq p_{\text {test}}$).
	As training and test data are distributed differently, models trained on training data are expected to generalize well to test data. 
	This calls for carefully designed OOD benchmarks to accurately assess models' generalization abilities.
	
	Taking inspiration from the recent fine-grained analysis of distribution shifts literature~\cite{wiles2021fine}, we provide a fine-grained analysis of distribution shifts on document images by dividing them into attributes related to image, text, and layout
	to investigate why a model $f$ trained on $p_{\text {train}}$ should generalize to $p_{\text {test}}$. 
	Specifically, a document image example is considered to be composed of the input $x$, label $y^l$, and its three attributes $\{y^{\text{image}}, y^{\text{text}}, y^{\text{layout}}\}$.
	As a convenience, we use $y^{1: K}$ to denote labels and attributes $\{y^l, y^{\text{image}}, y^{\text{text}}, y^{\text{layout}}\}$. 
	Then, we are able to formalize different distribution shifts associated with image, text, and layout for the generation of true data as follows:
	\begin{equation}
		p\left(y^{1: K}, \boldsymbol{x}\right)=p\left(y^{1: K}\right) p\left(\boldsymbol{x} \mid y^{1: K}\right)
	\end{equation}
	In this way, the data distribution can be expressed as the product of the marginal distributions of the decomposed attributes, which enables us to perform fine-grained analyses of various distribution shifts on document images.
	With the help of a latent variable model, the formalization can be written as follows:
	\begin{equation}
		p\left(y^{1: K}, \boldsymbol{x}\right)=p\left(y^{1: K}\right) \int p(\boldsymbol{x} \mid z) p\left(z \mid y^{1: K}\right) d z,
	\end{equation}
	where $z$ is the latent vector. 
	Through the above equation, different attributes $y^{1: K}$ can be used to affect latent variables $z$, thereby affecting the generation of data $\boldsymbol{x}$.

	\begin{figure*}[!htb]
		\centering
		\includegraphics[width=0.85\linewidth]{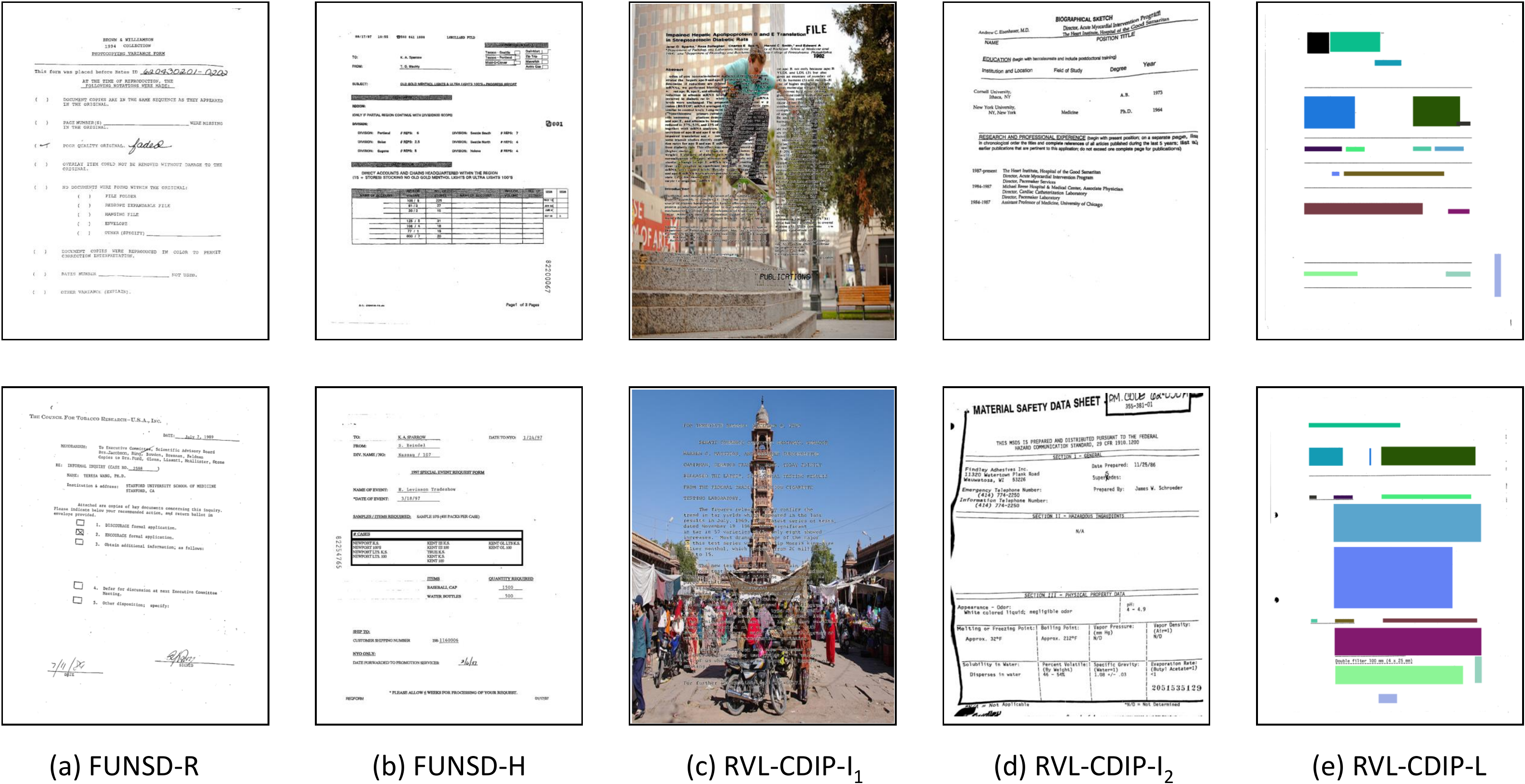}
		\caption{Samples of the distribution shift examples: (a) FUNSD-R containing a real-world OOD dataset variant of FUNSD, (b) FUNSD-H denoting the human-intervened OOD dataset variant of FUNSD, (c) RVL-CDIP-I$_1$ including samples with the natural image background, (d) RVL-CDIP-I$_2$ including samples with the distorted image background, and (e) RVL-CDIP-L containing samples with merged bounding boxes. More samples can be found at \color{gray}{\url{https://github.com/MAEHCM}}.
		}
		\label{fig:example}
	\end{figure*}

	\subsection{Image-Specific Distribution Shift}
	The natural and the distorted image background are two background variants for image distribution shifts.
	Formally, $y^{\text{image}}$ defines the image with the finite set $\mathcal{A} = \{a_{\text{original}}, a_{\text{natural}}, a_{\text{distorted}}\}$. 
	For training, the attribute $y^{\text{image}}$ is $a_{\text{original}}$. 
	During testing on out-of-distribution data with natural image backgrounds, we set the attribute $y^{\text{image}} = a_{\text{natural}}$ and then obtain marginal distribution over this attribute $p_{\text {natural }}(y^{1: K})$, which is used to induce the joint distribution over latent factors and the attribute \textit{natural}: 
	$p_{\text {natural }}(z, y^{1: K})=p(z \mid y^{1: K}) p_{\text {natural }}(y^{1: K})$. 
	Subsequently, we can get input data for testing with the joint distribution:
	$p_{\text {natural }}(\boldsymbol{x}, y^{1: K})$ equals to $\int p(\boldsymbol{x}\mid z) p_{\text {natural }}(z, y^{1: K})$. 
	On the other hand, the out-of-distribution test set with the background of distorted images $p_{\text {distorted }}(z, y^{1: K})$ can be derived in a similar way to the generation of test data with natural images.
	
	
	Practically, the new OOD benchmark with the joint distribution $p_{\text {natural }}(\boldsymbol{x}, y^{1: K})$ can be obtained through a two-stage pipeline: (1) \textbf{Disentangling text content from background}. We locate the text content based on the position information provided by an OCR tool and extract pixels of the text content from a document image. The rest of the pixels are viewed as background pixels; (2) \textbf{Replacing the original background with natural images}. We randomly select an image from MSCOCO and resize it to match the size of the document image. To compose a new OOD sample, the extracted text content is placed on the sampled natural image (Figure~\ref{fig:cases2} (b)).
	
	
	Document images may be distorted in real-world scenarios due to uncontrollable physical deformations, uneven illuminations, and various camera angles. To simulate this realistic environment, the OOD benchmark with the joint distribution $p_{\text {distort }}(\boldsymbol{x}, y^{1: K})$ is introduced. Inspired by~\cite{feng2022geometric}, we directly employ well-pretrained DocGeoNet to generate the OOD distorted images (Figure~\ref{fig:cases2} (a)).
	

	\subsection{Text-Specific Distribution Shift}
	To simulate a realistic scenario where input document images may contain problematic text caused by OCR errors, we employ two text attack strategies for text distribution shifts  (Figure~\ref{fig:cases2} (c)):
	(1) Bert-Attack;
	(2) Word Swap.
	Formally, $y^{\text{text}}$ defines the text with the finite set $\mathcal{A} = \{a_{\text{original}}, a_{\text{bert}}, a_{\text{swap}}\}$. 
	For training, the attribute $y^{\text{text}}$ is $a_{\text{original}}$. 
	Refer to the analysis of image-specific OOD benchmarks, the out-of-distribution test data with BERT-Attack $p_{\text {generation }}(\boldsymbol{x}, y^{1: K})$ and Word Swap $p_{\text {swap}}(\boldsymbol{x}, y^{1: K})$ can be obtained in a similar way.
	
	In practice, BERT-Attack, based on pre-trained masked language models exemplified by BERT, is used to produce OOD samples. The advantage of BERT-Attack is that it can generate similar but unseen words while guaranteeing fluency and semantic preservation in the generated samples. For Word Swap, we apply 5 ways to generate OOD samples: (1) \textit{Word Swap by embedding}: Using embedding vectors to find similar words to do swap; (2) \textit{Word Swap by homoglyph}: Replacing words with nearly identical in appearance yet different meaning; (3) \textit{Word Swap for numbers}: Replacing number with another number since numbers play an influential role in document images; (4) \textit{Random character deletion}: Deleting certain characters in words, such as "houses" $\longrightarrow$ "hoses".
	
	\subsection{Layout-Specific Distribution Shift}
	There are two layout manipulations for layout distribution shifts: Merge and Move. 
	The Merge manipulation is designed to investigate the impact of changing layout information from a fine-grained level to a coarse-grained level while maintaining image and text information.
	The move operation is used to investigate the effect of neighboring information on the content of a particular bounding box by moving the content to a distinct location. 
	The bounding box is an enclosed area surrounded by lines. 
	Formally, $y^{\text{layout}}$ defines the layout with the finite set $\mathcal{A} = \{a_{\text{original}}, a_{\text{merge}}, a_{\text{move}}\}$. 
	For training, the attribute $y^{\text{layout}}$ is $a_{\text{original}}$. 
	Refer to the analysis of image-specific and text-specific OOD benchmarks, the OOD test data with the Merge manipulation $p_{\text {merge }}(\boldsymbol{x}, y^{1: K})$ and the OOD test data with the Move manipulation $p_{\text {move}}(\boldsymbol{x}, y^{1: K})$ can be obtained in a similar way.
	
	The pseudocode of merging bounding boxes is shown in Algorithm \ref{alg:algorithm2}.
	To construct the OOD benchmark $p_{\text {merger }}(\boldsymbol{x}, y^{1: K})$, we perform the process described in Algorithm 1 for each image. 
	The Merge manipulation begins with initializing an empty set $\bs{S}$ which saves all the bounding boxes that have been traversed. 
	We then traverse each bounding box and get the current bounding box $\bs{B_i}$.
	Then it gets $\bs{B_{i}^{'}}$ by dilating $\bs{B_i}$ a litte using predefined horizontal and vertical dilation distances $\lambda_1$, $\lambda_2$.
	After that, we see if this dilated bounding box intersects with another bounding box in set $\bs{S}$.
	If there is an intersection, we get $\bs{M_i}$ by merging the two bounding boxes, otherwise, we skip this operation.
	We then add $\bs{M_i}$ or $\bs{B_{i}^{'}}$ to set $\bs{S}$.
	After all bounding boxes have been traversed, the merging process is complete  (Figure~\ref{fig:cases2} (e)) and we get merged bounding boxes $\bs{M_1}$,$\bs{M_2}$,...,$\bs{M_k}$.
	For the construction of OOD benchmark $p_{\text {move }}(\boldsymbol{x}, y^{1: K})$,
	we select a bounding box with strong textual semantics and then move the text content to another location without textual semantics (Figure~\ref{fig:cases2} (e)). 
	A semantic entity is considered to have strong text semantics if its prediction results remain unchanged after its corresponding layout information has been changed ten times.

	\begin{algorithm}[!htb]
		\caption{The procedure of merging bounding boxes.}
		\label{alg:algorithm2}
		\begin{algorithmic}[1]
			\REQUIRE    Bounding boxes $\bs{B_1}$,$\bs{B_2}$,...,$\bs{B_n}$ of a image; the collection $\bs{S}$ of bounding boxes that have been traversed; horizontal and vertical dilation distances $\lambda_1$, $\lambda_2$.
			\STATE		Initialize $\bs{S}$ to empty set.
			\REPEAT
			\STATE      Get the $i$-th bounding box $\bs{B_i} [x_1^i,y_1^i,x_2^i,y_2^i]$.
			\STATE      Dilate $\bs{B_i}$ with horizontal and vertical dilation distances as $\bs{B_{i}^{'}} [x_1^i-\lambda_1,y_1^i-\lambda_2,x_2^i+\lambda_1,y_2^i+\lambda_2]$.
			\IF{$\bs{B_{i}^{'}}$ intersects with a bounding box in the set $\bs{S_{i}}$ of $\bs{S}$}
			\STATE Merge the two bounding boxes in $\bs{S_{i}}$.
			\ENDIF
			\STATE		Mark the area where $\bs{B_{i}^{'}}$ is located in $\bs{S_{i}}$.
			\UNTIL      All the bounding boxes have been traversed.
			\ENSURE     Merged bounding boxes $\bs{M_1}$,$\bs{M_2}$,...,$\bs{M_k}$.
		\end{algorithmic}
	\end{algorithm}

	\section{Do-GOOD Datasets}
	The purpose of this section is to introduce the datasets used in our proposed Do-GOOD benchmark.
	we first perform a preliminary study for analysis of the distribution shift options for a specific VDU task.
	Then, we elaborate on OOD datasets across different VDU tasks.
	Finally, 9 datasets are constructed across 3 VDU tasks. 
	
	\subsection{Preliminary Study} 
	\label{sec:4.1}
	We conducted a preliminary study in order to investigate the effect of image, text, and layout information across different datasets and VDU tasks. Thus, for a certain dataset and task, we can determine which distribution shift should be chosen to develop the OOD test dataset.
	During implementation, we use LayoutLMv3$_{\text{BASE}}$~\cite{huang2022layoutlmv3} as the base model and isolate the effects of image, layout, and text information by removing the corresponding input embeddings for inference.
	Text is necessary for all tasks. Thus, to assess the effect of text information, we retain input text and remove image and layout embeddings.
	
	The overall results are shown in Table~\ref{tab:ablation_obj}.
	While the performance of LayoutLMv3 without (denoted as "w/o") image embeddings on RVL-CDIP drops substantially, the performance of information extraction slightly decreases and the performance of QA tasks may even increase.
	It indicates that document image classification is largely affected by image information. 
	We observe that without layout information, the performance of LayoutLMv3 drops by a significant margin for information extraction and classification. 
	However, model performance on DocVQA is not affected. 
	We assume that most questions in DocVQA are dependent on textual content to predict the answers.
	The performance of LayoutLMv3 only with text embeddings is slightly better than that with all input embeddings, which strongly supports our assumption of DocVQA. 
	Besides, using only text embeddings, LayoutLMv3 performs very poorly on information extraction and classification tasks. All of these analyses motivate us to develop image-specific OOD datasets for document image classification, text-specific OOD datasets for all tasks, and layout-specific OOD datasets for document image classification and information extraction.
	

	\begin{table}[!htb]
		\centering
		\caption{
			Overall results of the preliminary study on FUNSD, RVL-CDIP, and DocVQA datasets.
		}
			\begin{tabular}{lccc}
				\toprule			
				\multirow{2}{*}{\bf Model} & \bf FUNSD  & \bf RVL-CDIP & \bf DocVQA \\
				& \bf F1$\uparrow$ & \bf Accuracy$\uparrow$ & \bf ANLS$\uparrow$ \\
				
				\midrule
				$\textrm{LayoutLMv3}_{\rm BASE}$ & 90.29  & 95.44 & 78.76   \\
				\quad w/o Image & 90.18  & 57.07 & 78.82  \\
				\quad w/o Layout & 29.87  & 77.05 & 78.76 \\
				\quad w/ Text & 28.65  & 18.07 & 78.82    \\
				\bottomrule
			\end{tabular}
		\label{tab:ablation_obj}
	\end{table}

	\begin{table*}[!htb]
		\centering
		\caption{The ID and OOD performance of existing VDU models on the FUNSD, RVL-CDIP and DocVQA datasets. To compare the models fairly, all VDU models use cell-level layout embedding. Here $\textrm{OOD}_{\rm R}$ is an OOD dataset of FUNSD which samples 50 images from RVL-CDIP. $\textrm{OOD}_{\rm H}$ is our handcrafted dataset. $\textrm{OOD}_{\rm T}$, $\textrm{OOD}_{\rm L}$, $\textrm{OOD}_{\rm I_{1}}$ and $\textrm{OOD}_{\rm I_{2}}$ refer to text distribution shift, layout distribution shift, natural image distribution shift and distorted image distribution shift.
		}
		\begin{tabular}{l|ccccc|ccccc|cc}
			\toprule
			\multirow{2}{*}{\bf Model} &
			\multicolumn{5}{c}{\bf FUNSD} & 
			\multicolumn{5}{c}{\bf RVL-CDIP} &
			\multicolumn{2}{c}{\bf DocVQA}  
						\\ 
			& \bf ID & \bf $\textrm{OOD}_{\rm R}$ & \bf $\textrm{OOD}_{\rm H}$ & \bf $\textrm{OOD}_{\rm T}$ & \bf $\textrm{OOD}_{\rm L}$
			& \bf ID & \bf $\textrm{OOD}_{\rm T}$ & \bf $\textrm{OOD}_{\rm L}$ & \bf $\textrm{OOD}_{\rm I_{1}}$ 
			& \bf $\textrm{OOD}_{\rm I_{2}}$ &  \bf ID 
			& \bf $\textrm{OOD}_{\rm T}$  \\ 
			\midrule
			$\textrm{BROS}_{\rm BASE}$~\cite{hong2022bros} & 88.98 & 60.20 & \textbf{74.31} & 80.58 & 84.37 & 90.12 & 88.43 & 80.56 & 90.12  & 85.32 & 73.72 & 60.38   \\
			$\textrm{LiLT}_{\rm BASE}$~\cite{wang2022LiLT} & 88.25 & 57.45 & 72.32 & 78.41 & 68.83 & 95.68* & 85.31* & 51.20* & \textbf{95.68}* & 92.42* & 70.43 & 52.31  \\
			$\textrm{LayoutLM}_{\rm BASE}$~\cite{xu2020layoutlm} & 82.82 & 47.94 & 68.44 & 72.23 & 54.64 & 94.42 & 81.35 & 54.77 & 94.42 & 87.59 & 69.34 & 59.26  \\
			$\textrm{LayoutLMv2}_{\rm BASE}$~\cite{xu-etal-2021-layoutlmv2} & 89.91 & \textbf{62.39} & 72.70 & 79.16 & 81.33 & 95.25 & 86.53 &  64.78 & 82.08 & \textbf{92.16} & 78.08 & 64.67  \\
			$\textrm{LayoutLMv3}_{\rm BASE}$~\cite{huang2022layoutlmv3} & \textbf{90.29} & 57.88 & 73.25 & \textbf{86.82} & \textbf{84.95} & \textbf{95.44} & \textbf{89.32} & \textbf{81.06} & 86.27 & 85.02 & \textbf{78.76} & \textbf{65.69} \\
			\bottomrule
			\multicolumn{11}{l}{\footnotesize 
				* LiLT uses image features with ResNeXt101-FPN backbone in fine-tuning RVL-CDIP.
			}
		\end{tabular}
		\label{tab:compare_vdu}
	\end{table*}

	\subsection{Document Information Extraction Task}
	For the visual document information extraction task, we mainly generate OOD datasets based on FUNSD~\cite{jaume2019funsd}. FUNSD is a dataset sampled from the RVL-CDIP dataset~\cite{harley2015icdar} about noisy scanned form understanding, consisting of 199 documents (149 for training and 50 for testing) and 9,743 semantic entities. The task of FUNSD is sequential labeling, which aims to assign labels to words. 
	
	\textbf{FUNSD-L} is a variant of FUNSD that includes the OOD samples produced through strategies based on two layout-specific distribution shifts, Merger and Move. 
	As described in Section~\ref{sec:ben}, the Move operation is based on semantic strength determined by the model itself. 
	Specifically, we randomly shuffle the bounding boxes within a document image and employ the fine-tuned model to infer 30 times. 
	For textual content, if the model prediction has fewer errors, its semantic strength is greater. \textbf{FUNSD-T} is a variant of FUNSD that contains the OOD samples generated by the two text attack methods described in Section~\ref{sec:ben}.
	
	In fact, we also have OOD samples through observing and selecting from real-world datasets for the visual document information extraction task. Figure~\ref{fig:example} (c) and (d) show examples.
	Specifically, \textbf{FUNSD-R} is one real-world OOD dataset variant of FUNSD. 
	The FUNSD-R dataset is used to show the performance gap of VDU models on real-world OOD datasets and generated OOD datasets.
	First, we sample data examples from the large-scale document classification dataset RVL-CDIP, and then observe and select data examples that differ from the distribution of FUNSD. 
	After that, these selected examples are manually annotated. FUNSD-R contains 50 document images in total.
	Further, we modify the data examples in FUNSD in order to generate a human-intervented OOD dataset variant of FUNSD, named \textbf{FUNSD-H}. Practically, we move some weak textual entities to construct layout and image shift, or add a few semantically linked texts around strong semantic content to construct 3 kinds of shifts. In the end, we obtain 50 OOD samples.
	Despite the fact that it is expensive and time-consuming to construct OOD datasets such as FUNSD-R and RUNSD-H, these two OOD datasets inspired us to develop a suite of OOD benchmark datasets that can be generated automatically for a wide range of VDU tasks.
	
	\subsection{Visual Document Classification Task}
	RVL-CDIP~\cite{harley2015icdar} is a document classification dataset aiming to predict the category of a given document. It includes 400,000 data examples in 16 categories, which are divided into 320,000 training samples, 40,000 validation samples, and 40,000 test samples.
	
	\textbf{RVL-CDIP-T} is one of the OOD dataset variants of RVL-CDIP that contains the OOD samples generated by the two text attack methods. 
	As a variant of RVL-CDIP, \textbf{RVL-CDIP-L} includes OOD samples produced through two layout-specific distribution shifts. 
	The image-specific OOD variant, \textbf{RVL-CDIP-I}, is generated through natural \textbf{RVL-CDIP-I$_1$}and distorted \textbf{RVL-CDIP-I$_2$} image distribution shifts. 
	Examples are illustrated in Figure~\ref{fig:example}.
	
	\subsection{Document Visual Question Answering Task}
	DocVQA~\cite{mathew2021docvqa} is a dataset for predicting the answer given a document image and a question. To accomplish this, models need to understand the content of documents and learn to reason over them. The original DocVQA dataset consists of 10,194/1,286/1,287 images with 39,463/5,349/5,188 questions for training/validation/test, respectively.
	
	\textbf{DocVQA-T} is the OOD dataset variant of DocVQA. In order to construct DocVQA-T, we first collect text, questions, and answers from OCR results and the Microsoft READ API. Then, we obtain the OOD samples which are generated by the two text attack methods.

	\section{Experiment}
	\label{sec:exp}
	
	\subsection{Evaluation on state-of-the-art VDU Models}
	\noindent\textbf{VDU Models}. 
	Larger models are generally more robust to OOD data~\cite{hendrycks2020pretrained}.
	We thus evaluate the robustness of fine-tuning the popular pre-trained VDU models (large models) for downstream tasks on our Do-GOOD benchmark. 
	The state-of-the-art large models include (1) Pre-trained models with text and layout modalities: BROS~\cite{hong2022bros} and LiLT~\cite{wang2022LiLT}; (2) Pre-trained models with text, layout and image modalities: LayoutLMv1~\cite{xu2020layoutlm}, LayoutLMv2~\cite{xu-etal-2021-layoutlmv2}, and LayoutLMv3~\cite{huang2022layoutlmv3}.
	
	\noindent\textbf{Implementation Details}.
	We fine-tune the VDU models on the ID datasets while selecting the best checkpoints based on the performance of ID and OOD validation sets. 
	The evaluation metrics we use are the same as those used in the original dataset paper, such as ``\texttt{F1}'' for FUNSD and its OOD variants, ``\texttt{Accuracy}'' for RVL-CDIP and its OOD variants, and ``\texttt{ANLS}'' for DocVQA and its variants.
	All pre-trained models are based on Hugging Face~\cite{wolf-etal-2020-transformers}.
	For the visual document information extraction task, the learning rate is set to 3e-5, and the training epochs are set to 70. 
	Since the original RVL-CDIP corpus did not provide text information, we used the Tesseract 3 OCR engine to extract words and their positions. 
	The learning rate was set to 1e-6, and the training epoch was 30 rounds. For Doc VQA tasks, the learning rate is set to 2e-5, and the epoch is 40 rounds. 
	All input images have a resolution of 224 $\times$ 224 pixels, and the batch in training is set to 4, while the batch in testing is set to 1.
	
	\noindent\textbf{Main Results}.
	Based on the criteria outlined in Section 1, Do-GOOD is designed to achieve a large distribution gap between training and test data and a substantial performance drop from ID to OOD settings. 
	To verify whether the proposed OOD benchmark meets the criteria, we conduct experiments fine-tuning pre-trained VDU models on the original ID downstream datasets and testing on both the ID and OOD datasets. 
	
	Table~\ref{tab:compare_vdu} reports the overall results \textit{w.r.t} comparison of ID and OOD performance of the existing models on the FUNSD, RVL-CDIP and DocVQA datasets. 
	According to the differences between ID and OOD for each distribution shift across all VDU tasks, there is a substantial and consistent performance gap between the ID and OOD settings. 
	In most cases, LayoutLMv3 can achieve the best performance, including ID setting across all datasets, $\textrm{OOD}_{\rm T}$ and $\textrm{OOD}_{\rm I}$ settings of FUNSD and RVL-CDIP, and the $\textrm{OOD}_{\rm T}$ setting of DocVQA, indicating LayoutLMv3 is one of the most robust models on VDU tasks. These motivate us to use LayoutLMv3 as our base model for comparing common OOD algorithms on Do-GOOD benchmark. 
	BROS performs well in 4 OOD settings on FUNSD. The possible reason is that pre-training in BROS uses the relative position of the encoded text and a region masking strategy as the objective. 
	Based on the success of LayoutLMv3 and BROS, we assume that fine-grained modeling such as patch-level or region-level modeling may be very useful for improving the robustness of models in OOD environments. 
	
	\begin{table*}[!htb]
		\centering
		\caption{
			Overall comparison results of existing VDU models on the FUNSD and their own FUNSD-L datasets. FUNSD-L is generated by the move operation. Other Error, QA Error and Header Error refer to error rate of entities whose labels are other, question or answer, and header respectively.
		}
		\begin{tabular}{lccccccc}
			\toprule
			\multirow{2}{*}{\bf Model} & 
			\bf FUNSD & 
			\multicolumn{6}{|c}{\bf FUNSD-L}  \\
			& \bf F1$\uparrow$ & \multicolumn{1}{|c}{\bf Precision$\uparrow$} & \bf Recall$\uparrow$ & \bf F1$\uparrow$ & \bf Other Error$\downarrow$ 
			& \bf QA Error$\downarrow$ & \bf Header Error$\downarrow$  \\ 
			\midrule
			$\textrm{BROS}_{\rm BASE}$~\cite{hong2022bros} & 89.26 & \multicolumn{1}{|c}{77.13} & 93.10 & 84.37 & \textbf{43.97} & 1.79 & 100.00  \\
			$\textrm{LiLT}_{\rm BASE}$~\cite{wang2022LiLT} & 88.25 & \multicolumn{1}{|c}{60.03} & 80.66 & 68.83 & 54.92 & 13.18 & \textbf{59.46}  \\
			$\textrm{LayoutLM}_{\rm BASE}$~\cite{xu2020layoutlm} & 82.82 & \multicolumn{1}{|c}{53.68} & 55.64 & 54.64 & 82.88 & 42.32 & 72.73  \\
			$\textrm{LayoutLMv2}_{\rm BASE}$~\cite{xu-etal-2021-layoutlmv2} & 89.91 & \multicolumn{1}{|c}{71.74} & \textbf{93.89} & 81.33 & 81.73 & \textbf{1.24} & 95.74  \\
			$\textrm{LayoutLMv3}_{\rm BASE}$~\cite{huang2022layoutlmv3} & \textbf{90.29} & \multicolumn{1}{|c}{\textbf{80.40}} & 90.05 & \textbf{84.95} & 45.46 & 4.13 & 100.00 \\
			\bottomrule
		\end{tabular}
		\label{tab:compare_vdu_funsd}
	\end{table*}
	
	\begin{table*}[!htb]
		\centering
		\caption{Comparison results of existing VDU models under different text attack methods on the FUNSD dataset. All numerical results are averages of 5 runs. 
		}
		\begin{tabular}{lcccccc}
			\toprule
			\multirow{2}{*}{\bf Model} & 
			\bf Baseline & 
			\bf BERT-Attack & 
			\bf Embedding & 
			\bf Homoglyph & 
			\bf Change number &
			\bf Character deletion \\ & 
			\bf F1$\uparrow$ & \bf F1$\uparrow$ & \bf F1$\uparrow$ & 
			\bf F1$\uparrow$ & \bf F1$\uparrow$ & \bf F1$\uparrow$ \\ 
			\midrule
			$\textrm{BROS}_{\rm BASE}$~\cite{hong2022bros} & 88.98 & \textbf{89.04} & 82.55 & 66.56 & 89.23 & 75.51 \\
			$\textrm{LiLT}_{\rm BASE}$~\cite{wang2022LiLT} & 88.25 & 84.54 & 81.01 & 70.23 & 87.28 & 68.97 \\
			$\textrm{LayoutLM}_{\rm BASE}$~\cite{xu2020layoutlm} & 82.82 & 79.75 & 75.80 & 56.99 & 82.31 & 66.31 \\
			$\textrm{LayoutLMv2}_{\rm BASE}$~\cite{xu-etal-2021-layoutlmv2} & 89.91 & 86.61 & 83.60 & 60.83 & 89.27 & 75.53 \\
			$\textrm{LayoutLMv3}_{\rm BASE}$~\cite{huang2022layoutlmv3} & \textbf{90.29} & 88.14 & \textbf{86.44} & \textbf{84.50} & \textbf{90.10} & \textbf{84.91} \\
			\bottomrule
		\end{tabular}
		\label{tab:textoodres}
	\end{table*}
	
	\noindent\textbf{Results on FUNSD-L Dataset}.
	Furthermore, to explore the robustness of each model under the layout distribution shift condition, we evaluate the performance of each model in each label category on FUNSD-L.
	As shown in Table \ref{tab:compare_vdu_funsd}, we observe that LayoutLMv3 achieves the best performance on the FUNSD-L dataset. 
	BROS obtains the worst Other Error score. 
	LayoutLM \cite{xu2020layoutlm} and LayoutLMv2\cite{xu-etal-2021-layoutlmv2} have higher Other Error, indicating that the prediction of weak semantic areas can be easily affected by the layout of strong semantic areas in these models. 
	LayoutLM also has higher QA Error, indicating that the prediction of strong semantic entities may still be affected by the surrounding entities. 
	The low header accuracy of all models indicates that the prediction of the current model for the headers largely depends on the location of the entities.
	
	\noindent\textbf{Results on FUNSD-T Dataset}.
	Moreover, to investigate the robustness of each model under the text distribution shift condition, we evaluate the performance of the model against various text attacks.
	Table~\ref{tab:textoodres} shows the performance of each model under various attacks on FUNSD-T. We observe that LayoutLMv3\cite{huang2022layoutlmv3} achieves the best performance on 5 out of 6 text distribution shifts, which indicates that LayoutLMv3 is more robust than other models on text attacks.
	The performance gap between LayoutLMv3 and other VDU models on homoglyph is about 20 to 30 F1 score, which indicates that LayoutLMv3 model is more robust than other models when dealing with the semantic OOD caused by OCR error.

	\subsection{Evaluation on Typical OOD Algorithms}
	We further compare the representative OOD algorithms on all OOD datasets across three downstream tasks. 
	Based on the experiment results, we briefly analyze the effect of different OOD methods.
	All experimental results are based on $\textrm{LayoutLMv3}_{\rm BASE}$~\cite{huang2022layoutlmv3}.
	
	\noindent\textbf{Baseline Methods}.
	We use empirical risk minimization (ERM) and two OOD algorithms as our baselines. 
	ERM is a systematic process of identifying, assessing, and managing risks that face an organization. 
	The goal of ERM is to maximize the potential of positive events and minimize the impact of negative ones. 
	The 2 OOD methods are $\textrm{Deep Coral}$~\cite{sun2016deep} and $\textrm{Mixup}$~\cite{zhang2017mixup}. $\textrm{Deep Coral}$ achieves domain adaptive effects by aligning second-order statistics between the source and target domains.
	In our experiment, we only add this method at the last layer of $\textrm{LayoutLMv3}_{\rm BASE}$ model and set $\lambda$ equal to 1.
	$\textrm{Mixup}$~\cite{zhang2017mixup} achieves data augmentation without excessive overhead by interpolating input features and labels. 
	In our implementation, we simultaneously interpolate input features, including text embedding, layout embedding, and image embedding,  and then set $\alpha$ and $\beta$ equal to 0.4 for the Beta distribution.
	
	\begin{table*}[!htb]
		\centering
		\caption{The ID and OOD performances of 3 OOD algorithms on 12 datasets. All numerial results are averages of 5 runs.
		}
		\begin{tabular}{l|ccccc|ccccc|cc}
			\toprule
			\multirow{2}{*}{\bf Algorithm} &
			\multicolumn{5}{c}{\bf FUNSD} & 
			\multicolumn{5}{c}{\bf RVL-CDIP} &
			\multicolumn{2}{c}{\bf DocVQA}  \\ 
			& \bf ID & \bf $\textrm{OOD}_{\rm R}$ & \bf $\textrm{OOD}_{\rm H}$ & \bf $\textrm{OOD}_{\rm T}$ & \bf $\textrm{OOD}_{\rm L}$
			& \bf ID & \bf $\textrm{OOD}_{\rm T}$ & \bf $\textrm{OOD}_{\rm L}$ & \bf $\textrm{OOD}_{\rm I_{1}}$ 
			& \bf $\textrm{OOD}_{\rm I_{2}}$  & \bf ID & \bf $\textrm{OOD}_{\rm T}$  \\ 
			\midrule
			$\textrm{ERM}$ & \textbf{90.29} & 57.88 & 73.25 & \textbf{86.82} & \textbf{84.95} & \textbf{95.44} & 89.32 & \textbf{81.06} & 36.27 & 85.02 & \textbf{78.76} & \textbf{65.69}  \\
			$\textrm{Deep Coral}$~\cite{sun2016deep} & 90.20 & 58.88 & 73.92 & 84.61 & 83.47 & 95.12 & 89.21 & 76.57 & 37.82 & 86.23 & 78.63 & 64.21 \\
			$\textrm{Mixup}$~\cite{zhang2017mixup} &  89.28 & \textbf{61.19} & \textbf{74.33} & 86.53 & 84.23 & 94.69 & \textbf{89.87} & 78.70 & \textbf{40.44} & \textbf{87.09} & 77.66 & 65.34 \\
			\bottomrule
		\end{tabular}
		\label{tab:ood}
	\end{table*}

	\noindent\textbf{Main Results}.
	Table~\ref{tab:ood} shows the ID and OOD results of ERM, Deep Coral, and Mixup on 3 downstream tasks. 
	According to the observation in Table~\ref{tab:ood}, none of the OOD generalization algorithms consistently outperform ERM, and even ERM is superior to Deep Coral in most cases.
	Mixup can outperform ERM in $\textrm{OOD}_{\rm R}$ and $\textrm{OOD}_{\rm H}$ settings on FUNSD, but it underperforms ERM in $\textrm{OOD}_{\rm T}$ and $\textrm{OOD}_{\rm L}$ settings, indicating that developing a fine-grained comprehensive evaluation is of importance for OOD generalization.
	
	Next, we take a closer look at the effectiveness of the OOD algorithm on different tasks.
	For the information extraction task, Deep Coral and Mixup outperform ERM in $\textrm{OOD}_{\rm R}$ and $\textrm{OOD}_{\rm H}$. 
	These results demonstrate the rationality of our manual labeling data set and prove that common OOD algorithms are also applicable to VDU models. 
	In terms of layout distribution shifts on FUNSD, ERM performs slightly better than Deep Coral and Mixup. This may be due to the fact that common OOD algorithms are not capable of coping with excessive layout information changes.  
	For document image classification, Deep Coral and Mixup outperform ERM in $\textrm{OOD}_{\rm I_1}$ and $\textrm{OOD}_{\rm I_2}$ settings while they still perform worse than ERM in $\textrm{OOD}_{\rm L}$ settings.
	The results of this study indicate that common OOD algorithms perform well in the task of document classification, which requires a high level of image information modeling. 
	Deep Coral and Mixup score slightly below ERM on the document visual question answering task. The study demonstrates that distribution shifts in complex tasks such as document visual question answering cannot be easily handled using common OOD algorithms.
	
	\begin{figure}[!htb]
		\centering
		\includegraphics[width=0.65\linewidth]{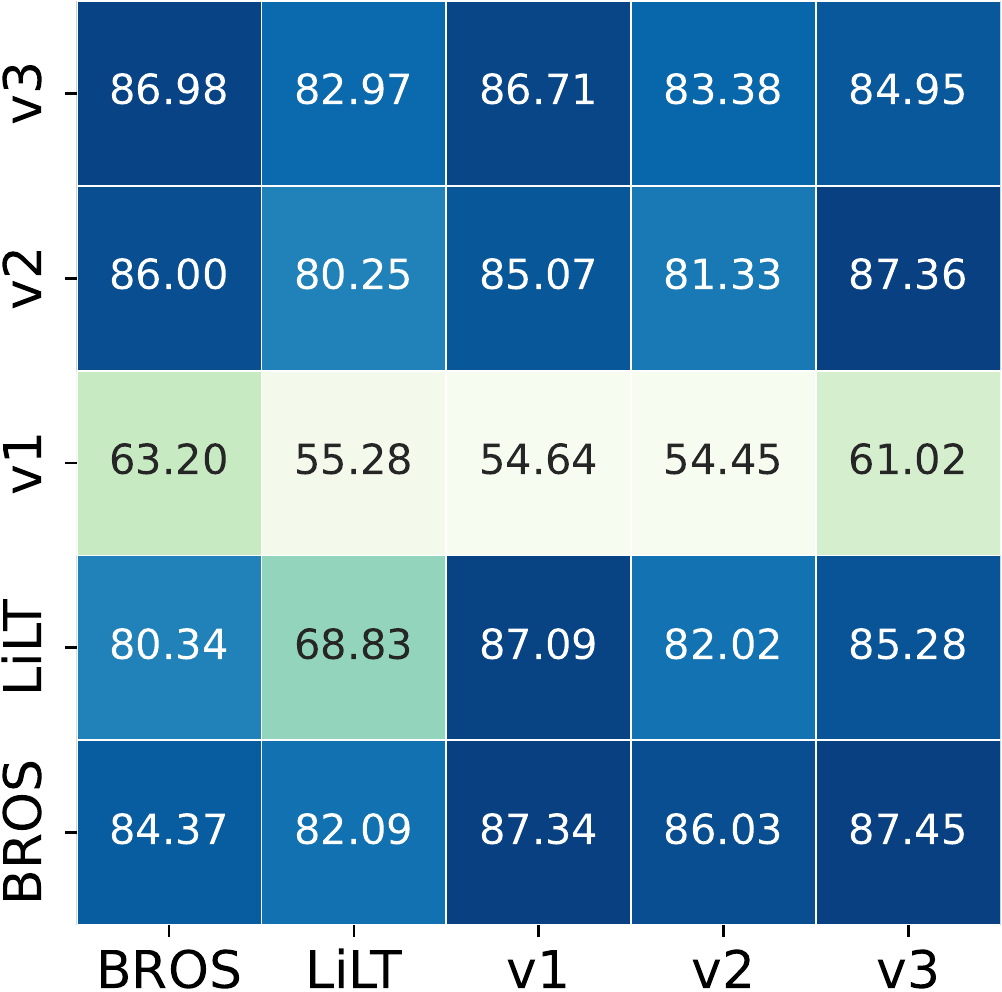}
		\caption{
			The confusion matrix in terms of F1 score for each VDU model on FUNSD-L data generated by the other models. The columns are the VDU models for generating the data, and the rows are the models for testing the data. v3 means LayoutLMv3, v2 means LayoutLMv2, v1 means LayoutLM.
		}
		\label{fig:confusion_matrix}
	\end{figure}
	
	\subsection{Further Analysis}
	\noindent\textbf{Effect of OOD Samples Generated by Different VDU Models.}
	As the generation of samples in FUNSD-L relies on the model to assess the semantic strength, we conduct experiments to investigate whether the performance of the model also drops substantially when OOD samples \textit{w.r.t} layout distribution shift are generated by other models.
	Figure~\ref{fig:confusion_matrix} shows the confusion matrix for each VDU model on FUNSD-L data generated by the other models. 
	We can observe that LayoutLM consistently performs worse on OOD datasets generated by all models. The results indicate that LayoutLM trained with fixed layout information is strongly dependent on layout information, which makes it difficult to cope with layout distribution shifts.
	Both LayoutLMv3 and BROS perform well on OOD datasets generated by all models, including themselves. 
	It demonstrates that fine-grained information modeling, such as patch-level and region-level information modeling, can improve the robustness of models.
	
	
	\begin{figure}[!htb]
		\centering
		\begin{subfigure}{0.236\textwidth}
			\centering
			\includegraphics[width=1.0\linewidth]{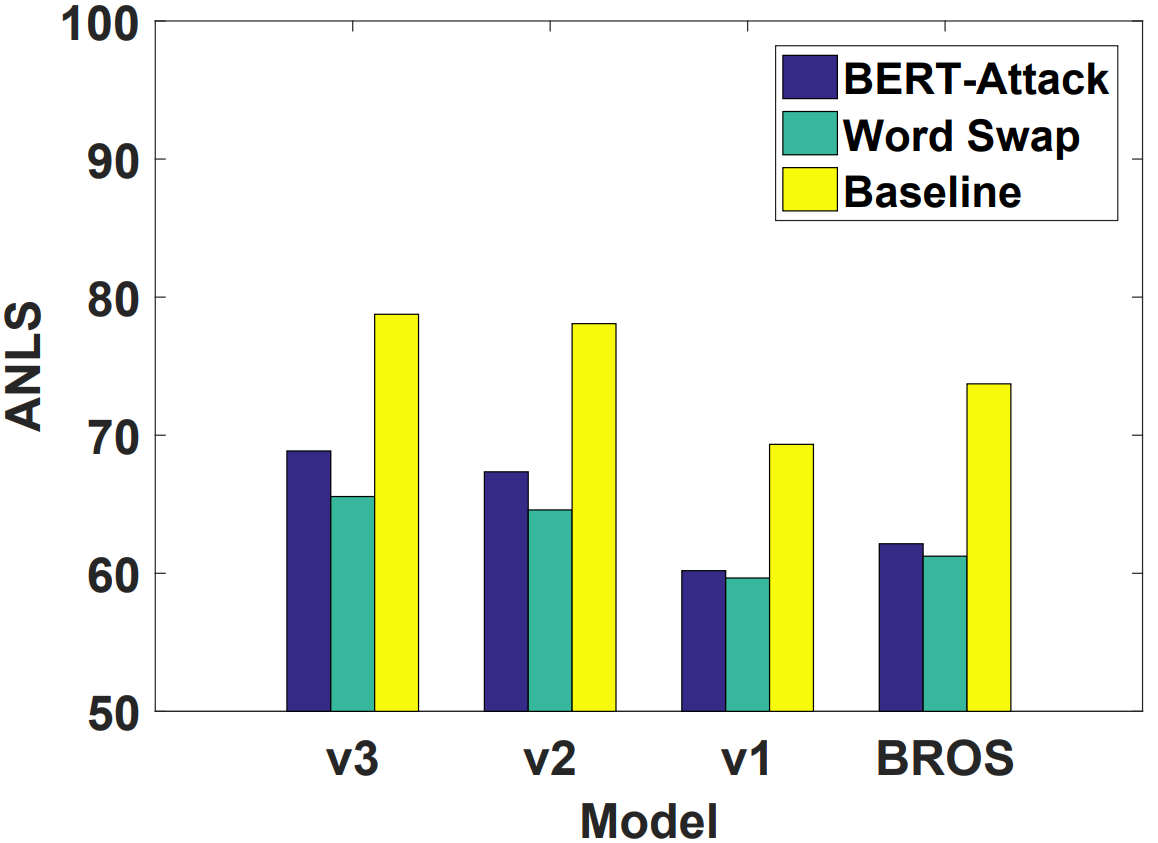}
			\caption{}
			\label{fig:ablation_vqatext}
		\end{subfigure}
		\begin{subfigure}{0.236\textwidth}
			\centering
			\includegraphics[width=1.0\textwidth]{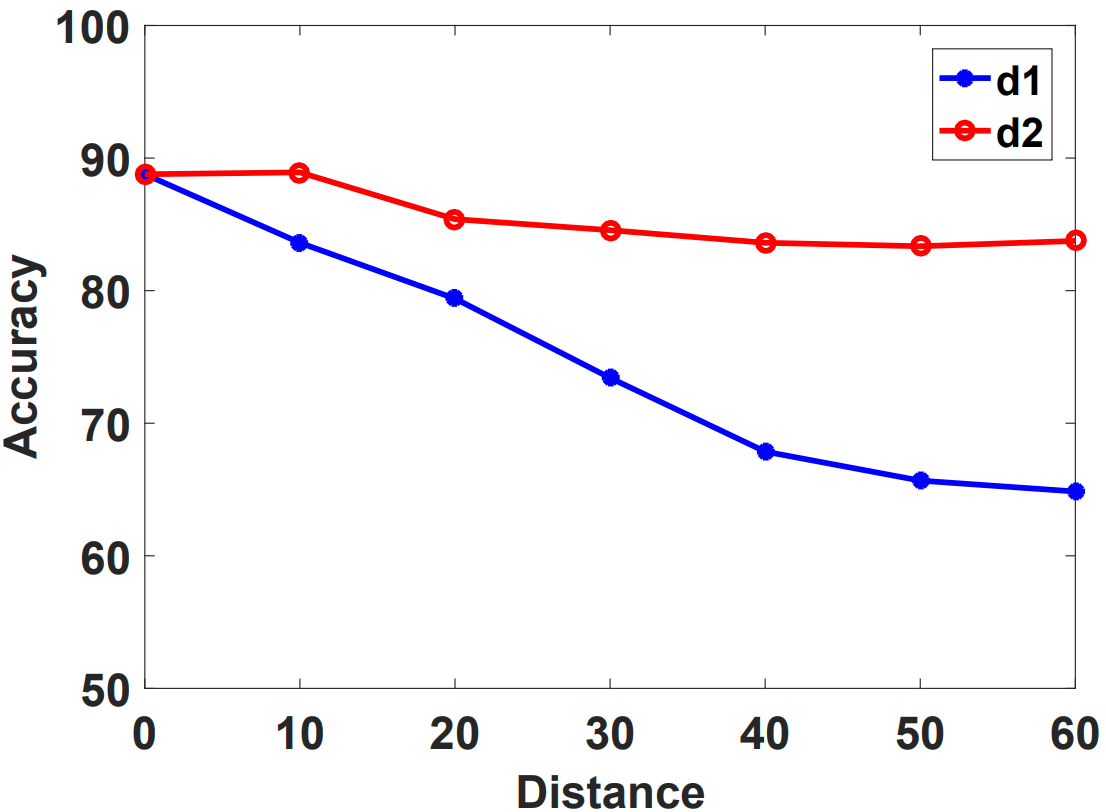}
			\caption{}
			\label{fig:ablation_layoutshift}
		\end{subfigure}
		\caption{Further analysis on (a) text distribution shift of VDU models on the DocVQA dataset. v3 means LayoutLMv3, v2 means LayoutLMv2, v1 means LayoutLM and (b) layout shift of LayoutLMv3 on document classification dataset RVL-CDIP.}
		\label{fig:ablation_curve}
	\end{figure}
	
	\noindent\textbf{Effect of Text Shift on Document VQA Task}.
	In Section \ref{sec:4.1}, we have demonstrated that LayoutLMv3 rarely uses the visual or layout information in document VQA tasks, thus for DocVQA test sets in this experiment, we only concentrate on the text information. 
	We utilize the same text shift method as FUNSD for text which is not answer. 
	Figure\ref{fig:ablation_vqatext} shows the results. 
	It can be seen that under the influence of Bert-Attack or Word Swap, the ANLS of all models dropped by about 10 points in terms of Accuracy. 
	It indicates that the existing VDU models are vulnerable to image corruption or OCR errors for document VQA task.
	
	\noindent\textbf{Effect of Merge Distance}.
	We further conduct experiments on the impact of distance parameter $d$, and the experimental results are shown in Figure~\ref{fig:ablation_layoutshift}. 
	Note that $d_{\rm 1}$ equals to $\lambda_{\rm 1}$ means vertical spacing and  $d_{\rm 2}$ equal to $\lambda_{\rm 2}$ means horizontal spacing.
	We can observe that some overlapping bounding boxes during OCR detection, thus, we explore whether the model needed fine-grained layout coordinates to predict document categories. Here $d_{\rm 1}$ controls the horizontal stretch length while $d_{\rm 2}$ controls the vertical stretch length. 
	When $d_{\rm 1}$ and $d_{\rm 2}$ are both 0, part of the OCR overlap area merges and the accuracy decreases. 
	It indicates that longitudinal merging reduces prediction accuracy more than horizontal merging. 
	
	\begin{figure}[!htb]
		\centering
		\begin{subfigure}{0.236\textwidth}
			\centering
			\includegraphics[width=0.87\linewidth]{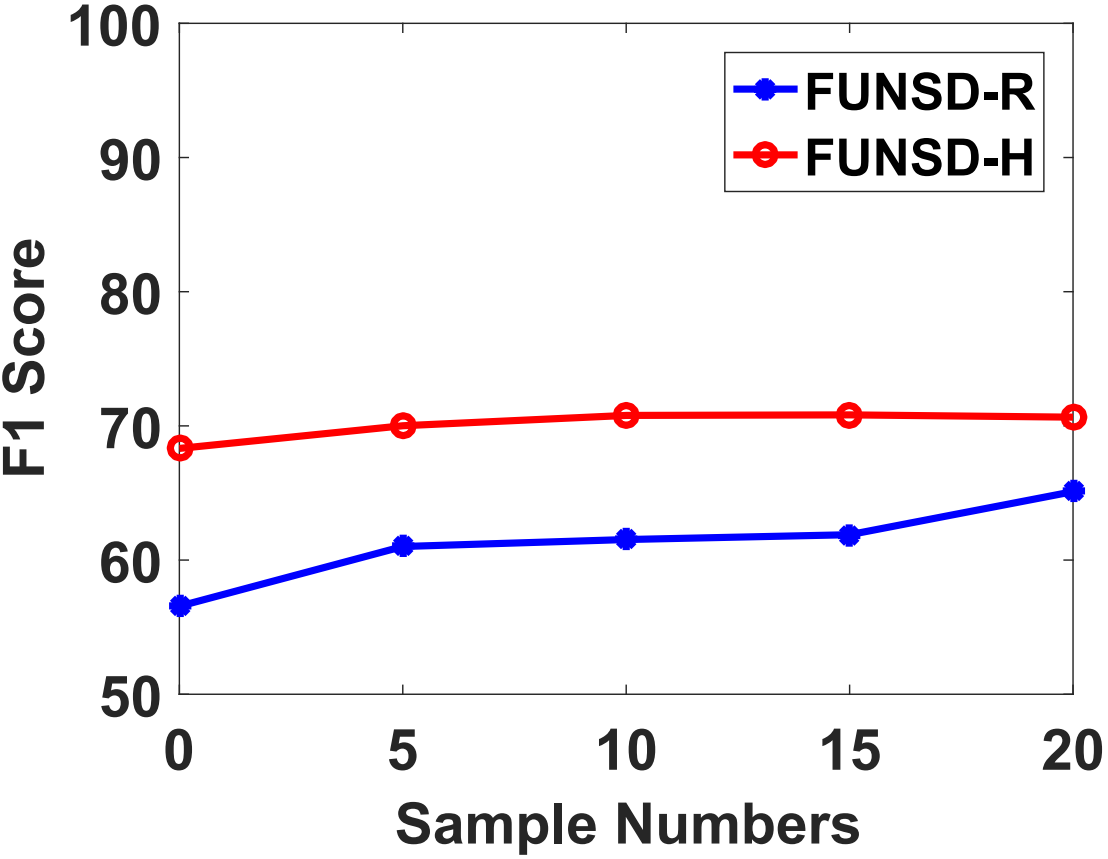}
			\caption{}
			\label{fig:line_chart_funsd}
		\end{subfigure}
		\begin{subfigure}{0.236\textwidth}
			\centering
			\includegraphics[width=0.9\textwidth]{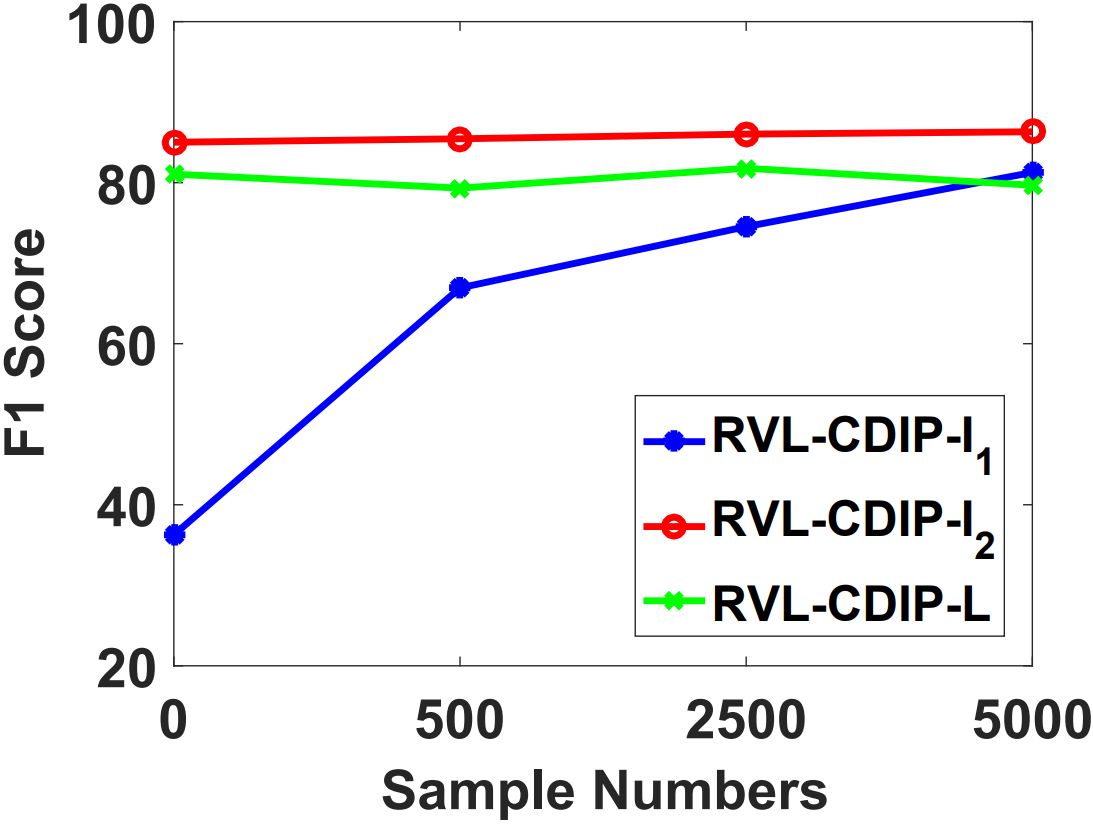}
			\caption{}
			\label{fig:line_chart_cdip}
		\end{subfigure}
		\caption{Incremental training results of LayoutLMv3 on (a) FUNSD-R and FUNSD-H, and (b)RVL-CDIP-$\rm I_1$, RVL-CDIP-$\rm I_2$ and RVL-CDIP-L datasets.}
		\label{fig:line_training}
	\end{figure}
	
	\noindent\textbf{Effect of Incremental Training with Do-GOOD}.
	We finally investigate the impact of the Do-GOOD benchmark on solving the OOD problem for existing VDU models considering the incremental training scheme. 
	Specifically, we divide the FUNSD-R and FUNSD-H datasets into training and test data split. Specially, 20 samples are added to the FUNSD training set for incremental training, 30 samples are tested as OOD samples. 
	We randomly sample five times and take the average of all results. 
	The experimental results show in Figure \ref{fig:line_chart_funsd} and we can see that adding OOD sample during training is effective to improve the performance of OOD test sets. 
	
	We further sample 5,000 samples on the RVL-CDIP validation set for incremental training and ensure that all document types are evenly distributed. 
	As shown in Figure \ref{fig:line_chart_cdip}, the experimental results show that adding OOD data to the training set can significantly improve the performance of the model on such OOD test sets when natural scene background replacement occurs for the document background. 
	For image distortion and Layout shift joining the training set to participate in incremental training, we find that the performance slightly changes on the OOD test set. 
	
	\section{Conclusion}
	\label{sec:conc}
	In this paper, we introduced an out-of-distribution (OOD) benchmark, \ies~ Do-GOOD, that evaluates the robustness of existing VDU models for document image-related tasks.
	We presented three criteria as well as a general, comprehensive framework for analyzing and benchmarking OOD document images.
	In this framework, we first broken down document images into image, text, and layout characteristics. 
	Then, we discussed the distribution shifts from image, text, and layout perspectives.
	We finally obtained 9 OOD datasets covering 3 document image-related tasks.
	On the basis of these OOD datasets, we conducted experiments using 5 existing pre-trained VDU models and two commonly used OOD generalization algorithms, which demonstrate the brittle nature of existing VDU models and OOD generalization algorithms.
	We expected that our framework and comprehensive benchmark will facilitate research in document image-related fields, and it can be utilized by practitioners to determine which methods perform best under which distribution shifts.

    \section{Acknowledgments}
This work was supported in part by National Natural Science Foundation of China under Grants (No. 62222203 and 61976049).
	\balance{
		\bibliographystyle{ACM-Reference-Format}
		\bibliography{bib_sigir2023_ood}
	}
	
\end{document}